\documentclass{article}
\usepackage{spconf,amsmath,graphicx}
\usepackage{algorithm}
\usepackage[noend]{algpseudocode}
\usepackage{amsfonts}
\usepackage{booktabs}

\newcommand{\tb}[1]{\textbf{#1}}
\title{Bayesian Neural Network For Personalized Federated Learning Parameter Selection}
\twoauthors
 {Mengen Luo}
	{Tsinghua University\\
	Tsinghua-Berkeley Shenzhen Institute\\
	ShenZhen, China}
 {Ercan Engin Kuruoglu}
	{Tsinghua University\\
	Tsinghua-Berkeley Shenzhen Institute\\
	ShenZhen, China}
\begin{document}
%
\maketitle
\begin{abstract}
Federated learning's poor performance in the presence of heterogeneous data remains one of the most pressing issues in the field. Personalized federated learning departs from the conventional paradigm in which all clients employ the same model, instead striving to discover an individualized model for each client to address the heterogeneity in the data. One of such approach involves personalizing specific layers of neural networks. However, prior endeavors have not provided a dependable rationale, and some have selected personalized layers that are entirely distinct and conflicting. In this work, we take a step further by proposing personalization at the elemental level, rather than the traditional layer-level personalization. To select personalized parameters, we introduce Bayesian neural networks and rely on the uncertainty they offer to guide our selection of personalized parameters. Finally, we validate our algorithm's efficacy on several real-world datasets, demonstrating that our proposed approach outperforms existing baselines.
\end{abstract}
\section{Introduction}
\label{sec:intro}

Recent artificial intelligence algorithms are predominantly data-driven, where the quantity of data plays a pivotal role in determining the effectiveness of these algorithms. However, the escalating awareness of security concerns and the increasingly stringent privacy regulations compel researchers and enterprises to seek methods that allow for training without compromising privacy. Federated learning emerges as a viable solution. It introduces a collaborative training approach without direct access to clients' raw data. Among the most favored algorithms within federated learning research is FedAvg\cite{mcmahan2017communication}. This method involves averaging client models after multiple rounds of local training to obtain a global model. Due to its efficiency in communication, FedAvg outperforms FedSGD, as it eliminates the need for communication in every round.

\begin{figure}
    \centering
    \centerline{\includegraphics[width=8.5cm]{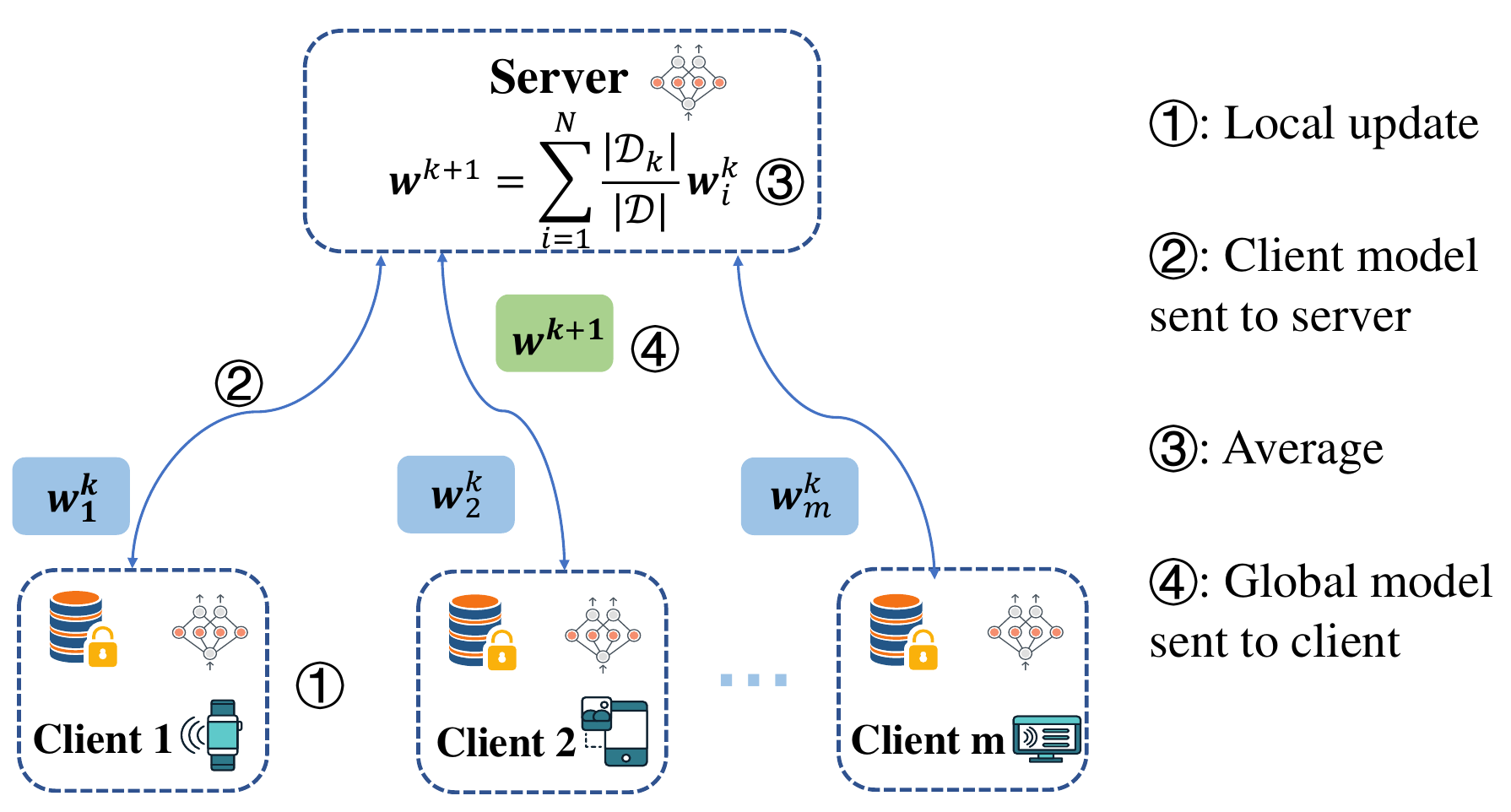}}
    \caption{FedAvg aggregates client models through averaging to obtain a global model, which is subsequently distributed to clients for training.}
    \label{fig:enter-label}
\end{figure}

However, recent experiments have unveiled the shortcomings of FedAvg. When client data distributions deviate from being independently and identically distributed (non-IID), the algorithm's performance experiences a sharp decline\cite{zhao2018fedwithnoniid}. To address this issue, several algorithms have sought to enhance FedAvg. FedProx employs a proximal term to constrain client training, ensuring that their models do not deviate excessively from the global model\cite{li2020fedprox}. Meanwhile, SCAFFOLD employs control variables to rectify client drift\cite{karimireddy2020scaffold}.

Another category of methods, known as Personalized Federated Learning (PFL), departs from some fundamental aspects of FedAvg. Instead of striving for a global optimum model, PFL seeks to provide each client with a personalized model to excel in their local data context. Given the complexity of real-world data and the diverse situations faced by individual clients, PFL aligns more closely with reality. A key consideration for a PFL algorithm lies in managing the interplay between shared knowledge and client-specific expertise. It must strike a balance between gaining advantages through collaborative training and effectively adapting to the unique challenges faced by each client.\

Prototype-based PFL aggregates client prototypes as common knowledge and utilizes global prototypes to guide client training\cite{tan2022fedproto}\cite{xu2023personalized}. Meanwhile, clustering-based algorithms posit that common knowledge exists among similar clients, aggregating clients within the same class\cite{sattler2020clustered}. Meta-learning-based algorithms employ a meta-learner as common knowledge to instruct client learning\cite{fallah2020personalized}\cite{acar2021debiasing}. The most prevalent approach involves personalization of model parameters, with a focus on layer-wise personalization. Algorithms such as FedPer \cite{arivazhagan2019federated} and LG-FedAVG \cite{liang2020think} personalize the model's feature extractor and classifier layers. These two algorithms have divergent opinions on which layers should be personalized, emphasizing the urgent need for an interpretable personalized parameters selection scheme.

\begin{figure}[htb]
  \centering
  \centerline{\includegraphics[width=8.5cm]{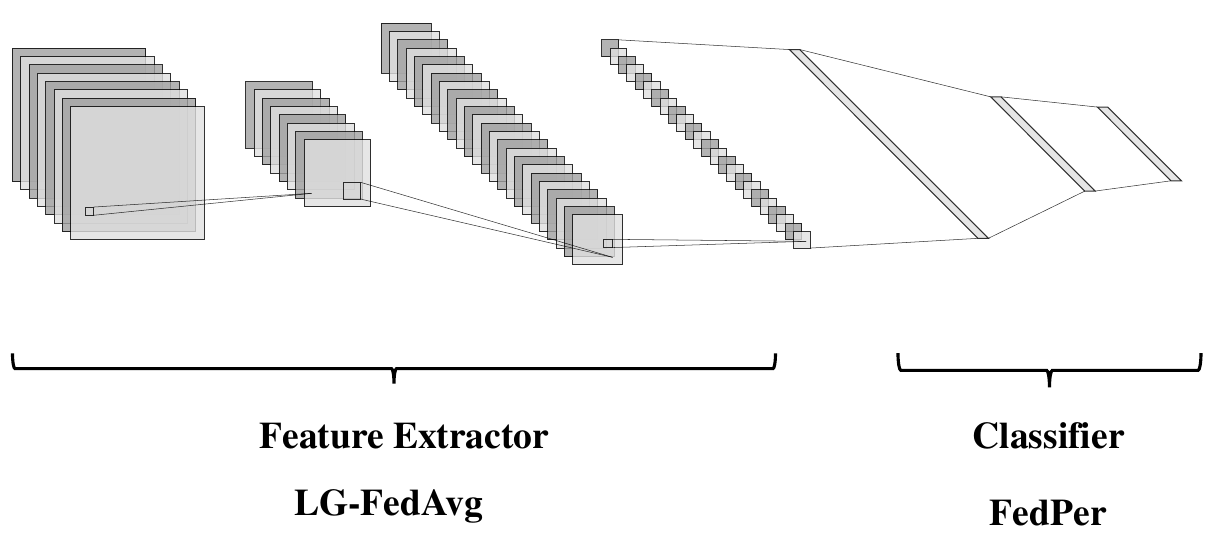}}

\caption{LG-FedAvg and FedPer diverge in their perspectives on which parts of the model should be personalized. LG-FedAvg posits that the feature extractor constitutes the personalized layer, whereas FedPer contends that the classifier serves as the personalized layer.}
\label{fig:res}
\end{figure}

To address this issue, we propose a simple yet effective approach that leverages the uncertainty of Bayesian neural networks to select personalized parameters. Parameters with higher uncertainty indicate a greater scope for improvement, with changes in these parameters exerting minimal impact on the final outcome compared to others. In the context of federated learning, the uncertainty of the global model also carries new significance. When employing model aggregation methods based on KL divergence, the global model's uncertainty encompasses the discrepancies among client models regarding the parameters. Consequently, parameters with higher uncertainty signify those where a consensus cannot be easily reached. We designate parameters with room for improvement and those exhibiting substantial client-to-client discrepancies as personalized parameters.

Through experiments conducted on various benchmark datasets, we have demonstrated the superiority of our method over FedPer and LG-FedAvg. This underscores the rationality of our approach to personalized parameters selection, which outperforms previous layer-based personalization techniques.

\section{Problem Setting}
\label{sec:problem}

Suppose we have $N$ clients and a central server. In personalized FL, each client has its distinct dataset, denoted as $\mathcal{D}_{1}, \ldots, \mathcal{D}_{N}$. Here, $\mathcal{D}_{i}={(x_{n}, y_{n})}_{n=1}^{N{i}}$ represents the labeled data of client $i$. Our model is denoted as $f_{w}\colon \mathcal{X}\to \mathcal{Y}$, where $w \in \mathbb{W}$ represents the model parameters, and we employ the loss function $\ell\colon \mathcal{X} \times \mathcal{Y} \to \mathbb{R}$. There is a mask $M$ have same shape with $w$, represent the personalized parameters of $w$, then the underlying optimization goal of PFL is:

\begin{equation}
    \min\limits_{W}\{F(W) := \frac{1}{N} \sum_{k=1}^{N} \mathbb{E}[\ell(w_g, w_i;x,y)]\}
\end{equation}
where $W=(w_1,w_2,...,w_n)$ denotes the collection of all client models and $w_g$ denotes global model. The goal of element level parameter personalization is to identify a mask $M$ that determines which parameters should be personalized. The loss can be rewrite as $\mathbb{E}[\ell(w_i\odot M + w_g \odot (1-M);x,y)]$.

\section{Proposed Method}
\label{sec:proposed}

In this section, we will discuss the method we propose, namely Federated BNN for Parameter Selection (FedBPS). We employ Bayesian neural networks to guide our selection of personalized parameters. Bayesian neural networks assume a distribution of parameters. In our case, We assume no correlation between elements; hence, the parameter distribution of the neural network will follow a diagonal Gaussian distribution. This distribution can be characterized by its mean and variance, where higher variance indicates that the parameters have a larger adjustment range, making them suitable candidates for personalized parameters.

\subsection{Bayesian Neural Network}
The key distinction between Bayesian neural networks and point-estimate neural networks lies in how Bayesian neural networks treat parameters as distributions rather than a single fixed values. Consequently, Bayesian neural networks provide a measure of uncertainty regarding network parameters. When a parameter exhibits a high variance, it implies that the parameter's value is uncertain, and changes in this parameter have a smaller impact on model performance compared to other parameters. Therefore, such parameters are well-suited for use as personalized parameters, as they can accommodate personalized tasks without significantly affecting the global subnetwork performance.


Bayesian neural network posterior estimation employs two main approaches: Variational Inference (VI)\cite{graves2011practical}\cite{blundell2015weight}\cite{dusenberry2020efficient} and Markov Chain Monte Carlo (MCMC)\cite{welling2011bayesian}\cite{zhang2020cyclical}\cite{el2021federated}. The MCMC involves sampling from the posterior to estimate its density, while VI makes parametric assumptions about the posterior and seeks to find the distribution within a family of distributions that is closest to the posterior distribution by minimizing the Evidence Lower Bound (ELBO). Both of these methods entail higher computational and storage overhead compared to point-estimate neural networks.

To acquire the posterior distribution of parameters, we employ the Laplace approximation method\cite{daxberger2021laplace}.  Since we do not require precise posterior results but rather their relative magnitudes, the Laplace approximation aligns perfectly with our needs. Laplace approximation utilizes second-order derivative information at a specific point to approximate the distribution as a Gaussian distribution, with the distribution depicted as follows:

\begin{equation}
\label{eq:laplace}
    w \sim \mathcal{N}(w_{0}, (\nabla_{w}^2 \mathcal{L}(\mathcal{D};w)|_{w_{0}})^{-1})
\end{equation}

Compared to algorithms based on MCMC and VI, Laplace approximation is faster and can be seamlessly integrated with point-estimate neural network algorithms since it is a post hoc method.

\subsection{Bayesian Neural Network For Personalized Parameters Selection}

In this section, we will discuss how to select personalized parameters based on uncertainty. Let's assume we have a matrix $W$ representing the parameters, where $n_W$ denotes the total number of elements in $W$. Let $p$ denote the proportion of personalized parameters we require, so we need $pn_W$ personalized parameters. We create a matrix $M$ of the same shape as $W$, where its elements are composed of $0$ and $1$. A threshold $a$ make there are $pn_W$ elements in variance $\sigma$ bigger than $a$.

\begin{equation}
\label{eq:mask}
    M_{ij}=
    \left\{
    \begin{array}{cc}
        1 & \text{if } \sigma \geq a\\
        0 & \text{if } \sigma \leq a
    \end{array}
    \right.
\end{equation}

Thus, personalized parameters can be represented as:

\begin{equation}
    w_i = w_i \odot M + w_g \odot (1- M)
\end{equation}
where $\odot$ is Hadamard product and $w_g$ is the global model. This formula signifies that any new parameter $w_i$ will originate from either $w_i$ or $w_g$. When the parameter's variance is low, it will adopt the value of the global model parameter, and when the variance is high, this personalized parameters will adopt the value of $w_i$. 

When dealing with a single model, parameter selection based on uncertainty is straightforward. However, in the context of personalized federated learning, each client possesses their own personalized model. Consequently, selecting personalized parameters can lead to disparities. To address this, We will first aggregate the global parameter distribution $\mathcal{N}(\mu_g,\sigma_g)$ and then determine which parameters should undergo personalization by using $\sigma_g$. In this context, we opt for an aggregation scheme based on KL divergence.

\begin{equation}
\label{eq:bnn aggregation}
    \begin{aligned}
        \mu_g &= \sum_{i=1}^{N} \pi_i \mu_i\\
        \sigma_g &= \sum_{i=1}^{N} \pi_i \sigma_i + \pi_i(\mu_i- \mu)^2\\
    \end{aligned}
\end{equation}


From the formula, it is evident that the variance of the aggregated model is determined by the mean of client model variances and the variance of that mean. Consequently, the global model comprises components with low variance and minimal discrepancies among clients, while personalized parameters exhibit high variance or substantial inter-client disparities.

\begin{algorithm}[tb]
\caption{Personalized Federated BNN for Parameter Selection}
\label{alg:pfedpf}
\textbf{Input}: Datasets $\mathcal{D}_i$, initial NN weight $w_i$\\
\textbf{Output}: $w_i, i=1,\ldots,n$
\begin{algorithmic}[1] 
\Procedure{Server Executes}{}
    \For{$t=1,2, \ldots, T$}
        \For{$i=1,2,\ldots, N$}
            \State $w_i,\sigma_i \leftarrow$ \Call{ClientUpdate}{$w_i$}
        \EndFor
        \State $w_g = \sum_{i=1}^{N} \pi_i w_i$
        \State $\sigma_g = \sum_{i=1}^{N} \pi_i \sigma_i + \pi_i(w_i - w)$
        \State Get threshold $a$ using proportion $p$
        \State Calculate mask $M$ from $\sigma$ by Equation \ref{eq:mask}
        \State $w_i = w_i \odot M + w_g \odot (1- M)$
    \EndFor
    
\EndProcedure

\Procedure{ClientUpdate}{w}
    \For{$e=1,2, \ldots, E$}
        \State $w \leftarrow w - \eta\nabla f(w)$
    \EndFor
    \State $\sigma \leftarrow \nabla_{w}^2 \mathcal{L}(\mathcal{D}_i;w)$
    \State \textbf{Return} $w,\sigma$
\EndProcedure

\end{algorithmic}
\end{algorithm}

\section{Experiments}
\label{sec:experiment}

\subsection{Setup}

\textbf{Datasets} We evaluate our algorithm and the baseline on three popular datasets: MNIST, which consists of handwritten digits with 10 classes; Fashion-MNIST, featuring 10 clothing categories; and CIFAR-10, containing colored images with 10 classes.

\textbf{Data Partition} To simulate each client's heterogeneous dataset, we follow the approach of \cite{karimireddy2020scaffold}\cite{xu2023personalized}. Each client possesses an equal number of samples, but they are divided into major and minor classes. We set the number of major classes to 2, with the remaining classes designated as minor. The proportion of major classes to the total client data is represented as $s=0.2$ which quantifies the degree of data heterogeneity. In line with the personalized federated learning setup, we apply the same partitioning to the test set to ensure that both the training and test sets are drawn from the same distribution.

\begin{table}[t]
\centering
\caption{The final test accuracy on different datasets under non-IID settings.}
\begin{tabular}{llll}
\toprule
Method &
MNIST&
FMNIST &
CIFAR10\\
\midrule
FedAvg & \tb{98.49} & 87.28 & 71.53\\
FedPer & 96.48 & 92.09 & 75.27 \\
LG-FedAvg & 96.44& 91.92 & 76.52\\
\midrule
FedBPS(ours) & 96.87 & \tb{92.51} & \tb{76.80}\\
\bottomrule
\end{tabular}
\label{tab:accuracy}
\end{table}

\begin{table}[t]
\centering
\caption{The impact of different datasets on hyperparameters in a heterogeneous data setting.}
\begin{tabular}{llll}
\toprule
Proportion &
MNIST&
FMNIST &
CIFAR10\\
\midrule
30\% & 95.44 & 91.13 & 75.27\\
50\% & \tb{97.74} & \tb{93.14} & 75.72 \\
70\% & 96.87 & 92.52 & \tb{76.80}\\
90\% & 96.23 & 92.19 & 76.23\\
\bottomrule
\end{tabular}
\label{tab:hp}
\end{table}

\textbf{Baselines} We will compare the following three baselines: FedAvg, representing the traditional setup where the goal is to learn a global model; FedPer, which employs the classifier as a personalized layer; and LG-FedAvg, which uses the feature extractor as a personalized layer. By contrasting FedPer and LG-FedAvg with these layer-based personalization algorithms, we aim to gain intuitive insights into the effectiveness of the algorithms. Our method, FedBPS, using $70\%$ of the parameters as personalized parameters.

\textbf{Implementation details} For each method, we will adopt common settings for local training. We will use SGD as the local optimizer with a learning rate of $\eta=0.01$, a batch size of $B=128$, weight decay set to 5e-4, and momentum set to 0.9. In the context of federated learning, the local epoch will be set to $E=5$. The total communication rounds will be set to $T=60$ for MNIST and Fashion-MNIST, and $T=100$ for CIFAR-10. For the neural networks, we will employ LeNet-5\cite{lecun1998gradient} for MNIST and Fashion-MNIST. For CIFAR-10, we will make adjustments to LeNet by preserving its architectural structure while increasing the network's width, all networks with two sets of convolutional and pooling layers, followed by three fully-connected layers. To investigate the effect of network depth on hyperparameters, we will conduct experiments using ResNet\cite{he2016resnet} architectures of varying depths on the CIFAR-10 dataset.


\subsection{Numerical Results}
We conducted experiments on three distinct datasets, and it is evident from the results in Table \ref{tab:accuracy} that our method outperforms the baseline in most scenarios. The only exception is observed in the case of the MNIST dataset, where FedAvg exhibits superior performance. Nevertheless, our approach still achieves higher accuracy compared to layer-level personalized methods. These experiments substantiate the effectiveness of our algorithm.

\begin{figure}
    \centering
    \includegraphics[width=8.5cm]{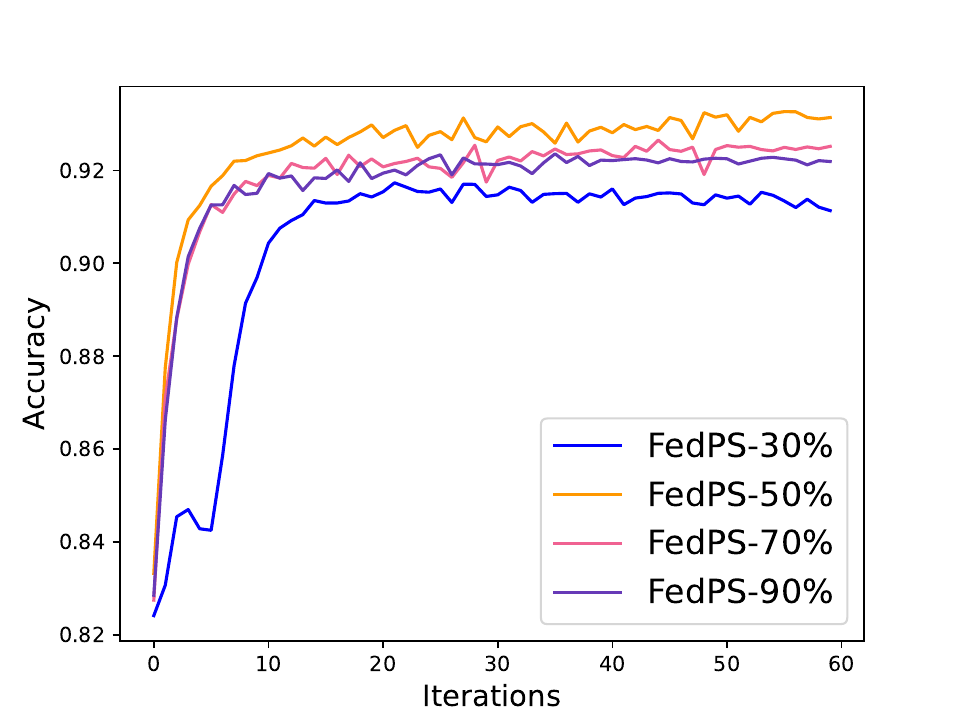}
    \caption{The impact of personalization proportion on accuracy under the FMNIST dataset.}
    \label{fig:hp_acc}
\end{figure}

Clearly, the proportion of personalized parameters has a profound impact on the algorithm's effectiveness. Here, we continue to assess the influence of hyperparameters on experimental results across these three datasets. We conducted experiments with personalization proportions of $30\%, 50\%, 70\%, 90\%$, and the final outcomes are presented in Table \ref{tab:hp}. It is noteworthy that the optimal personalization proportion varies significantly depending on the dataset and model used. These results can only suggest that the optimal proportion may fall between $50\%$ and $90\%$. 

For ResNet architectures of different depths, the result are presented in Table \ref{tab:resnet}. The recommended personalization proportion falls within the range of 70\% to 100\%, which differs from the earlier LeNet architecture. However, it is evident that the depth of network with similar architecture has minimal influence on the choice of personalization proportion.

\begin{table}[t]
\centering
\caption{The impact of the depth of ResNet on hyperparameters in a heterogeneous data setting.}
\begin{tabular}{lllll}
\toprule
 &
ResNet20&
ResNet32 &
ResNet44 &
ResNet56 \\
\midrule
30\% & 72.13 & 72.47 & 72.17 & 71.75\\
50\% & 73.21 & 72.96 & 73.19 & 73.34\\
70\% & 74.92 & 74.25 & 73.94 & 73.73\\
90\% & 75.08 & 74.96 & 74.49 & 74.60\\
\bottomrule
\end{tabular}
\label{tab:resnet}
\end{table}

\section{Conclusion}
\label{sec:conclusion}

In this paper, we introduced an approach for selecting personalized parameters to enable personalized federated learning, leveraging the uncertainty obtained from Bayesian neural networks for parameter selection. Our method offers enhanced flexibility, allowing for the selection of personalized parameters at a finer granularity, and incurs additional costs well within acceptable bounds compare with other Bayesian approach. Through experiments conducted on various datasets, we have validated the superior performance of our approach compared to layer-level personalized federated learning.

\vfill\pagebreak

\bibliographystyle{IEEEbib}
\bibliography{refs}

\begin{thebibliography}{10}

\bibitem{mcmahan2017communication}
Brendan McMahan, Eider Moore, Daniel Ramage, Seth Hampson, and Blaise~Aguera
  y~Arcas,
\newblock ``Communication-efficient learning of deep networks from
  decentralized data,''
\newblock in {\em Artificial Intelligence and Statistics}. PMLR, 2017, pp.
  1273--1282.

\bibitem{zhao2018fedwithnoniid}
Yue Zhao, Meng Li, Liangzhen Lai, Naveen Suda, Damon Civin, and Vikas Chandra,
\newblock ``Federated learning with non-iid data,''
\newblock {\em arXiv preprint arXiv:1806.00582}, 2018.

\bibitem{li2020fedprox}
Tian Li, Anit~Kumar Sahu, Manzil Zaheer, Maziar Sanjabi, Ameet Talwalkar, and
  Virginia Smith,
\newblock ``Federated optimization in heterogeneous networks,''
\newblock {\em Proceedings of Machine Learning and Systems}, vol. 2, pp.
  429--450, 2020.

\bibitem{karimireddy2020scaffold}
Sai~Praneeth Karimireddy, Satyen Kale, Mehryar Mohri, Sashank Reddi, Sebastian
  Stich, and Ananda~Theertha Suresh,
\newblock ``Scaffold: Stochastic controlled averaging for federated learning,''
\newblock in {\em International Conference on Machine Learning}. PMLR, 2020,
  pp. 5132--5143.

\bibitem{tan2022fedproto}
Yue Tan, Guodong Long, Lu~Liu, Tianyi Zhou, Qinghua Lu, Jing Jiang, and Chengqi
  Zhang,
\newblock ``Fedproto: Federated prototype learning across heterogeneous
  clients,''
\newblock in {\em Proceedings of the AAAI Conference on Artificial
  Intelligence}, 2022, vol.~36, pp. 8432--8440.

\bibitem{xu2023personalized}
Jian Xu, Xinyi Tong, and Shao-Lun Huang,
\newblock ``Personalized federated learning with feature alignment and
  classifier collaboration,''
\newblock {\em International Conference on Learning Representations}, 2023.

\bibitem{sattler2020clustered}
Felix Sattler, Klaus-Robert M{\"u}ller, and Wojciech Samek,
\newblock ``Clustered federated learning: Model-agnostic distributed multitask
  optimization under privacy constraints,''
\newblock {\em IEEE Transactions on Neural Networks and Learning Systems}, vol.
  32, no. 8, pp. 3710--3722, 2020.

\bibitem{fallah2020personalized}
Alireza Fallah, Aryan Mokhtari, and Asuman Ozdaglar,
\newblock ``Personalized federated learning with theoretical guarantees: A
  model-agnostic meta-learning approach,''
\newblock {\em Advances in Neural Information Processing Systems}, vol. 33, pp.
  3557--3568, 2020.

\bibitem{acar2021debiasing}
Durmus Alp~Emre Acar, Yue Zhao, Ruizhao Zhu, Ramon Matas, Matthew Mattina, Paul
  Whatmough, and Venkatesh Saligrama,
\newblock ``Debiasing model updates for improving personalized federated
  training,''
\newblock in {\em International Conference on Machine Learning}. PMLR, 2021,
  pp. 21--31.

\bibitem{arivazhagan2019federated}
Manoj~Ghuhan Arivazhagan, Vinay Aggarwal, Aaditya~Kumar Singh, and Sunav
  Choudhary,
\newblock ``Federated learning with personalization layers,''
\newblock {\em arXiv preprint arXiv:1912.00818}, 2019.

\bibitem{liang2020think}
Paul~Pu Liang, Terrance Liu, Liu Ziyin, Nicholas~B Allen, Randy~P Auerbach,
  David Brent, Ruslan Salakhutdinov, and Louis-Philippe Morency,
\newblock ``Think locally, act globally: Federated learning with local and
  global representations,''
\newblock {\em arXiv preprint arXiv:2001.01523}, 2020.

\bibitem{graves2011practical}
Alex Graves,
\newblock ``Practical variational inference for neural networks,''
\newblock {\em Advances in Neural Information Processing Systems}, vol. 24,
  2011.

\bibitem{blundell2015weight}
Charles Blundell, Julien Cornebise, Koray Kavukcuoglu, and Daan Wierstra,
\newblock ``Weight uncertainty in neural network,''
\newblock in {\em International Conference on Machine Learning}. PMLR, 2015,
  pp. 1613--1622.

\bibitem{dusenberry2020efficient}
Michael Dusenberry, Ghassen Jerfel, Yeming Wen, Yian Ma, Jasper Snoek,
  Katherine Heller, Balaji Lakshminarayanan, and Dustin Tran,
\newblock ``Efficient and scalable bayesian neural nets with rank-1 factors,''
\newblock in {\em International Conference on Machine Learning}. PMLR, 2020,
  pp. 2782--2792.

\bibitem{welling2011bayesian}
Max Welling and Yee~W Teh,
\newblock ``Bayesian learning via stochastic gradient langevin dynamics,''
\newblock in {\em Proceedings of the 28th International Conference on Machine
  Learning}, 2011, pp. 681--688.

\bibitem{zhang2020cyclical}
Ruqi Zhang, Chunyuan Li, Jianyi Zhang, Changyou Chen, and Andrew~Gordon Wilson,
\newblock ``Cyclical stochastic gradient mcmc for bayesian deep learning,''
\newblock {\em International Conference on Learning Representations}, 2020.

\bibitem{el2021federated}
Khaoula El~Mekkaoui, Diego Mesquita, Paul Blomstedt, and Samuel Kaski,
\newblock ``Federated stochastic gradient langevin dynamics,''
\newblock in {\em Uncertainty in Artificial Intelligence}. PMLR, 2021, pp.
  1703--1712.

\bibitem{daxberger2021laplace}
Erik Daxberger, Agustinus Kristiadi, Alexander Immer, Runa Eschenhagen,
  Matthias Bauer, and Philipp Hennig,
\newblock ``Laplace redux-effortless bayesian deep learning,''
\newblock {\em Advances in Neural Information Processing Systems}, vol. 34, pp.
  20089--20103, 2021.

\bibitem{lecun1998gradient}
Yann LeCun, L{\'e}on Bottou, Yoshua Bengio, and Patrick Haffner,
\newblock ``Gradient-based learning applied to document recognition,''
\newblock {\em Proceedings of the IEEE}, vol. 86, no. 11, pp. 2278--2324, 1998.

\bibitem{he2016resnet}
Kaiming He, Xiangyu Zhang, Shaoqing Ren, and Jian Sun,
\newblock ``Deep residual learning for image recognition,''
\newblock in {\em Proceedings of the IEEE Conference on Computer Vision and
  Pattern Recognition}, 2016, pp. 770--778.

\end{thebibliography}

\end{document}